\def\year{2022}\relax
\documentclass[letterpaper]{article} 
\usepackage{aaai22}  
\usepackage{times}  
\usepackage{helvet}  
\usepackage{courier}  
\usepackage[hyphens]{url}  
\usepackage{graphicx} 
\usepackage{natbib}  
\usepackage{caption} 
\DeclareCaptionStyle{ruled}{labelfont=normalfont,labelsep=colon,strut=off} 
\title{A Factor-Based Framework for Decision-Making Competency Self-Assessment}
\author{
    Brett W. Israelsen,\textsuperscript{\rm 1}
    Nisar Ahmed, \textsuperscript{\rm 2}
}
\affiliations{
    \textsuperscript{\rm 1} Artificial Intelligence, Raytheon Technologies Research Center\\
    \textsuperscript{\rm 2} Smead Aerospace Engineering Sciences, University of Colorado Boulder\\
    brett.israelsen@rtx.com, nisar.ahmed@colorado.edu
}

\usepackage{amsmath,amsthm,amssymb,amsfonts}
\usepackage{bm}

\newcommand{\famsec}{FaMSeC}

\newcommand{\xQ}{\ensuremath{x_Q}} 
\newcommand{\xO}{\ensuremath{x_O}} 
\newcommand{\xP}{\ensuremath{x_P}} 
\newcommand{\xI}{\ensuremath{x_I}} 
\newcommand{\xM}{\ensuremath{x_M}} 
\newcommand{\xind}{\ensuremath{x_i}} 

\def\-{\raisebox{.75pt}{-}} 
\newcommand{\ri}{\ensuremath{R_{\infty}}}
\newcommand{\riref}{\ensuremath{\bar{R}_{\infty}}}

\newcommand{\piopt}{\ensuremath{\pi^*}}

\newcommand{\lwr}[1]{\ensuremath{L_{#1}}}
\newcommand{\upr}[1]{\ensuremath{U_{#1}}}
\newcommand{\flow}{\ensuremath{\lwr{\xind}}}
\newcommand{\fup}{\ensuremath{\upr{\xind}}}

\usepackage{xcolor}

\begin{document}

\maketitle


\section{Introduction}
Assessing an autonomous robot's fitness for a task requires human supervisors or users to know how well-suited the underlying `ingredients of autonomy' are for meeting their expectations, in spite of imperfect and uncertain information available at both design time and deploy time \cite{Muir1994-ow, Israelsen2019-survey}. Since such stakeholders are not expected to be experts in such matters, robots must close this loop in a readily accessible way---much as delegated autonomous human subordinates are expected to interact with their supervisors \cite{Miller2014-av}. 
It is thus important for robots to be able to identify the limits of their \emph{competency} \cite{Hutchins2015,Israelsen2020-xQ}, i.e. proficiency \cite{Gautam2021}. We address the challenge of doing this in a general algorithmic way. 

As an analogy: consider how a correctly tuned Kalman filter can be trusted to `know' its true state estimation error statistics through the computed state error covariance matrix, such that the assumed approximate filtering distribution model is statistically validated by truth model simulations and correctly fits actual sensor data \cite{Bar-Shalom2001}. A robot with a correct sense of `algorithmic self-trust' could also be similarly trusted to assess its own limitations on a task --- provided the self-trust measure truly reflects how well the robot's underlying functionality (with all its attendant models, data, assumptions, algorithms, and other approximations of reality) fits the task. 
Given that many algorithmic approaches for decision-making under uncertainty are based on probabilistic models and reasoning methods \cite{Kochenderfer2015}, we argue that these quantitatively rich components provide ample opportunity to rigorously extend this analogy and develop concrete self-trust assessment methods for broad problem classes. Specifically, such models and methods admit readily computable and interpretable stochastic meta-analyses for determining where they are expected to break down/remain valid and exceed/deceed desired task performance specs. 

We summarize our efforts to date in developing these ideas to build a framework for generating 
succinct human-understandable competency self-assessments in terms of \emph{machine self-confidence}, i.e. a robot's self-trust in its functional abilities to accomplish assigned tasks. Whereas early work explored machine self-confidence in ad hoc ways for niche applications \cite{Hutchins2015, Kuter2015,Sweet2016, Zagorecki2015, Kaipa2015-hy}, our \emph{Factorized Machine Self-Confidence (\famsec{})} framework introduces and combines several aspects of probabilistic meta-reasoning for algorithmic planning and decision-making under uncertainty to arrive at a novel set of generalizable \emph{self-confidence factors}, which can support competency assessment for a wide variety of problems. 

\section{Factorized Machine Self-Confidence}

\begin{figure}[t]
\centering
\includegraphics[width=0.99\columnwidth]{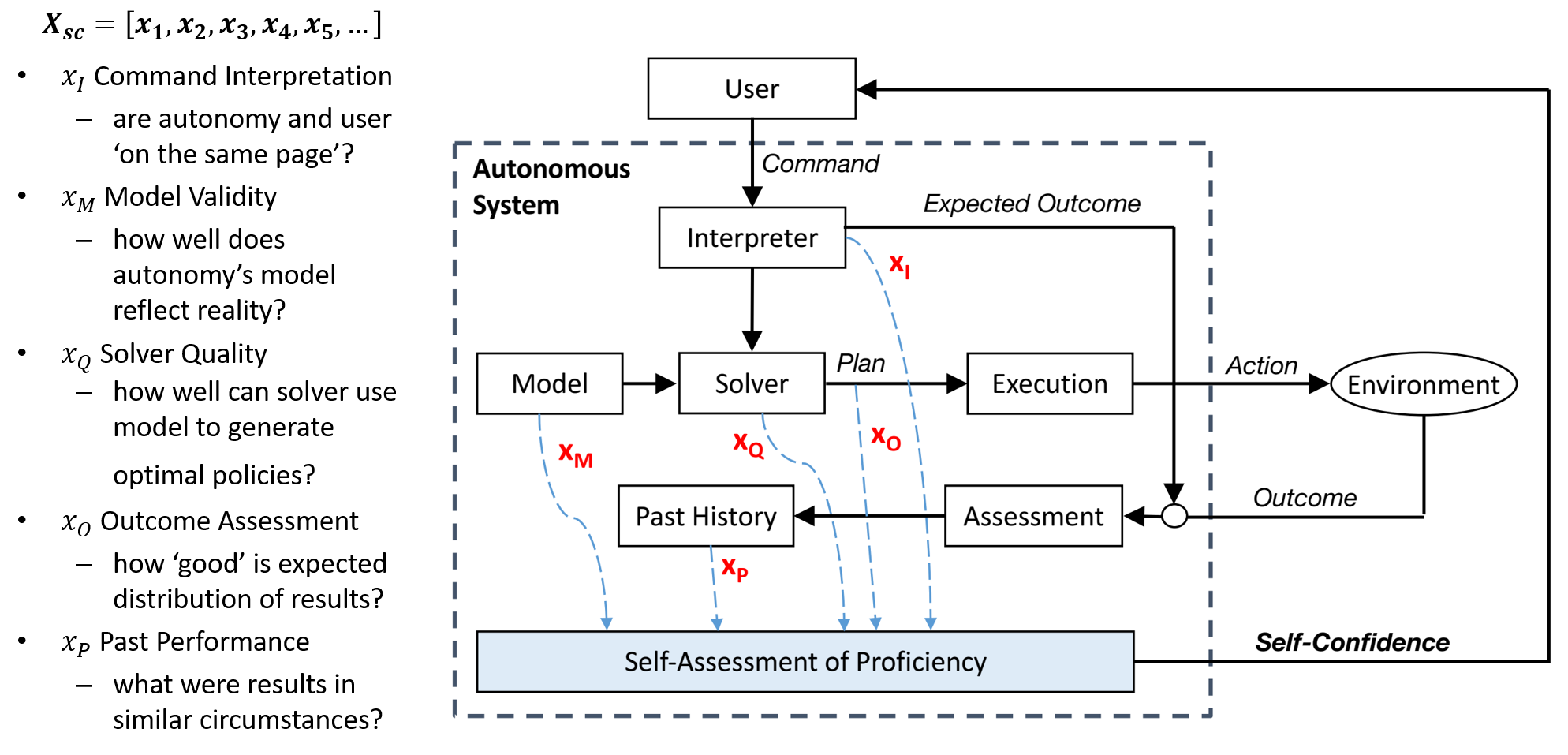}
\caption{\famsec{} overview: self-confidence factors ($\xind{}$  dashed lines) are derived from core algorithmic decision-making components (white boxes in the `Autonomy' block). }
\label{fig:famsec}
\end{figure}

As shown in Fig. \ref{fig:famsec}, \famsec{} represents and computes self-confidence factors through functions that score different parts of an autonomous planning agent's decision-making process. The combined set of self-confidence factors provide a meta-analysis of how operations and approximations inherent to the agent's decision-making process are expected to impact it's overall task performance. 
As with the self-confidence reporting strategy developed in \cite{Hutchins2015}, the factors ideally encode an expert designer's assessment of the reliability and suitability of an autonomous decision-making system for completing a task, accounting for variations and uncertainties in task specification, context, environment, and system implementation. \famsec{}'s novelty is that it allows autonomous decision-makers to automatically generate self-confidence assessments from information already available from setting up/solving the task at hand. Thus, unlike \cite{Hutchins2015}, experts are not needed to assign self-confidence values a priori for every possible scenario that the decision-maker could encounter. 
Each factor \xind{} can be mapped onto an arbitrary scale, e.g. \flow{} to \fup{} for the sake of discussion, where \flow{} gives a shorthand indication of `complete lack of confidence' (i.e. some aspect of task falls completely outside competency boundaries), and \fup{} indicates `complete confidence' (i.e. all aspects of task are well within competency boundaries). 
The scales for each factor need not all be the same 
as long as a clear `confidence directions' 
can be established for each. 


\subsection{Summary of Results to Date and Ongoing Work}
\citet{Aitken2016} considers five general self-confidence factors for algorithmic planning and decision-making (Fig. \ref{fig:famsec} left), which could also be combined to produce a total self-confidence score that acts as a simple shorthand signal for users and then parsed further as needed. 
These five factors are not necessarily exhaustive and are primarily aimed at meta-analysis of decision-making \emph{prior} to the execution of a particular task, e.g. other self-confidence factors could be included to account for in situ and post hoc analyses.  
%
To date, two `downstream' \famsec{} factors of the five -- Outcome Assessment (\xO{}) and Solver Quality (\xQ{})  -- have been studied, under the assumption that `upstream' factors are fixed at `ideal' confidence levels (e.g. as if user intent perfectly understood, models perfectly describe task, etc.). \xO{} assesses the inherent difficulty of a specified task instance relative to the available plan (or policy) produced by the decision-making algorithm instance (solver) and known set of user expectations. 
\xQ{} assesses the ability of the given decision-making algorithm instance to arrive at a solution that completely satisfies known task objectives. 

An approach for computing \xO{} was developed in \citet{Aitken2016} for infinite horizon MDPs. It relies on comparing the achievable cumulative \emph{actual} reward value $\ri= \sum_{k=0}^{\infty}R_{k}$ earned via following an optimal policy \piopt (where $R_k$ is the task reward at time step $k$). The approach quantifies how much probability mass lies to the right vs. left of a user-specified reference minimally acceptable cumulative reward value \riref{}, which defines a lower-bound on acceptable task performance expectations, e.g. $\riref=0$ means any reward greater than 0 is acceptable. The ratio of expected exceedance to deceedance relative to \riref{} forms the basis for \xO{}-based competency assessment, i.e. the degree to which $\ri$ exceeds (deceeds) \riref{} with a positive (negative) margin directly translates to higher (lower) task confidence. This is captured by the upper partial moment/lower partial moment ratio statistic (UPM/LPM) \cite{kong2006upm}.  
Since the UPM/LPM may be unbounded, it is also logistically transformed, scaled, and shifted to some desired $[L_{x_O},U_{x_O}]$ range, e.g. $[-1,+1]$, to facilitate interpretability.   
    \begin{figure}[tbp]
        \centering
        \includegraphics[width=0.95\linewidth]{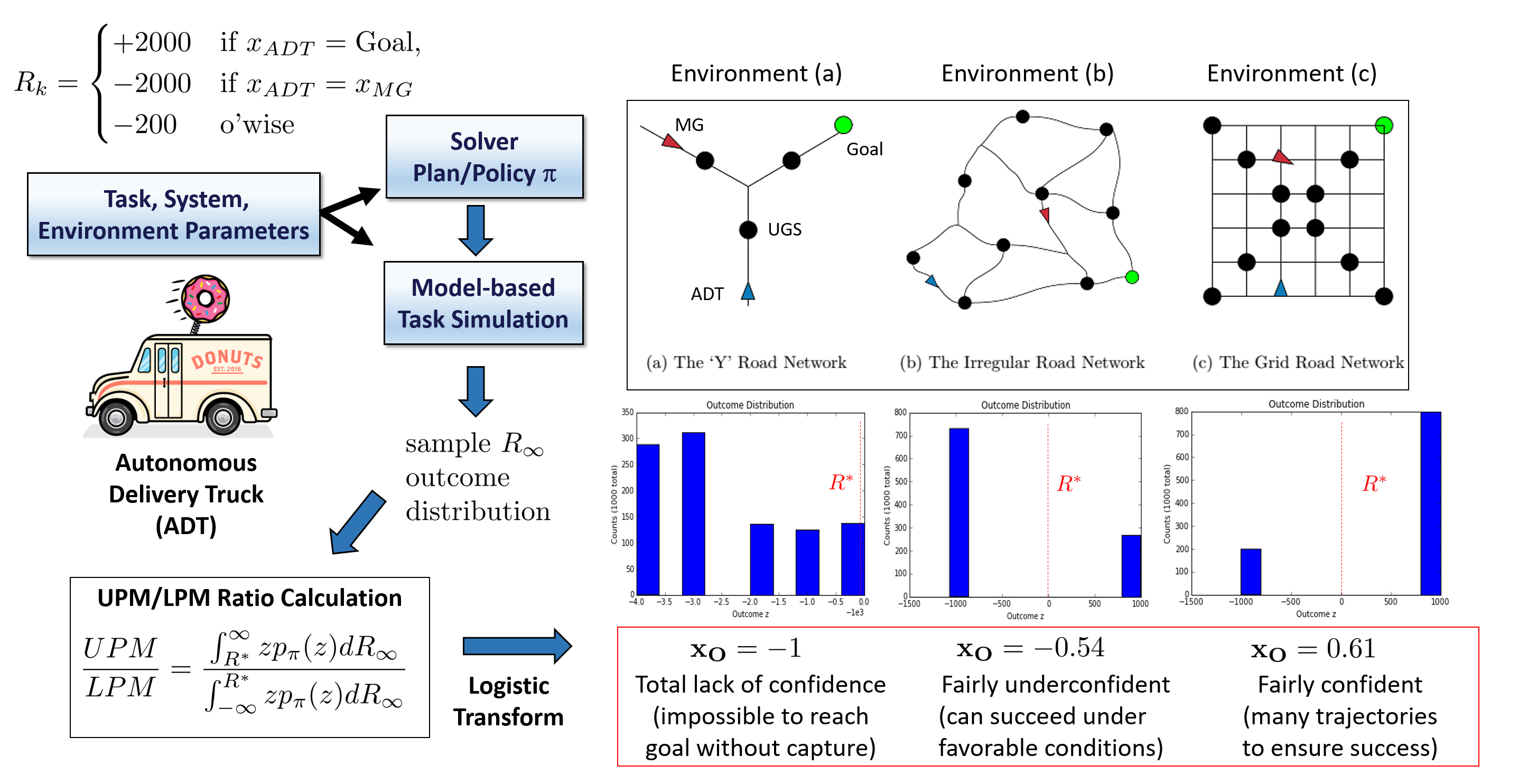}
        \caption{\xO{} assessments for MDP planning instances where an autonomous delivery truck (ADT) must navigate to goal and avoid motorcycle gang (MG) \cite{Israelsen2020-xQ}.}
        \label{fig:xOexample}
    \end{figure}
\noindent Fig.~\ref{fig:xOexample} illustrates the behavior of \xO{} in three different example MDP-based value iteration planning scenarios. In `Environment (a)', all of the probability mass in $p(\ri)$ is left of \riref{}, resulting in $\xO=-1$. In (b) and (c), \xO{} changes sign due to higher probability mass being left or right of \riref{}, indicating how well a task matches demanded competency limits. 

A method for assessing \xQ{} was developed in \citet{Israelsen2020-xQ}. Inspired by learning techniques for empirical hardness modeling \cite{Leyton-Brown2009}, it relies on learning a surrogate model for figures of merit of `trusted' solver (i.e. an offline `kitchen sink' or gold-standard method) on a given task in different known environment and task configurations, and then using this model to compare to figures of merit obtained online for a deployed (likely sub-optimal and resource-constrained) `candidate' solver in an arbitrary task setting. The comparison may be performed using any surrogate model learning technique (Gaussian processes, deep neural nets, etc.) and any easily computable figures of merit for determining the `quality' of a planning solution, e.g. expected cumulative rewards or total reward distributions at solution convergence. As with \xO{}, higher/lower \xQ{} confidence scores for a given task instance correspond to the degree to which the candidate solver's figures of merit exceeds/deceeds those predicted by the trusted solver's surrogate model. 
%


Using a simulation of the delivery task in Fig. \ref{fig:xOexample}, Israelsen \cite{Israelsen2019-thesis} conducted a large ($N=255$ person) study of compensated MTurk users to assess whether \xO{} and \xQ{} reports (provided as Likert values mapped from numerical scores) could improve supervisory performance (measured by total delivery score and time spent on multiple tasking instances) and self-reported trust (measured by follow up surveys). Users had to decide only whether to dispatch the truck or not on 40 randomly assigned tasking instances, based on self-confidence reports. 
Using a between subjects design for the different self-confidence reporting cases, analysis of the data revealed that the presence of either/both FaMSeC factors had strong positive effect on cumulative score, a weaker positive impact on self-reported trust, and a negligible impact on decision making time. Besides developing computable metrics for the remaining factors (\xM{}, \xI{}, \xP{}), 
ongoing work includes: (i) extending $\xO{}$ to both non-reward based generalized outcome assessments formulated in terms of task-relevant outcome semantics and to learning-based `black-box' world models (e.g. deep neural nets) \cite{ConlonSciTech2022}; (ii) impact of competency report communication in relation to different problem scenarios; and (iii) adaptation to perception and other reasoning problems. 
%

\bibliography{mybib}

\begin{thebibliography}{17}
\providecommand{\natexlab}[1]{#1}

\bibitem[{Aitken et~al.(2016)Aitken, Ahmed, Lawrence, Argrow, and
  Frew}]{Aitken2016}
Aitken, M.; Ahmed, N.; Lawrence, D.; Argrow, B.; and Frew, E. 2016.
\newblock Assurances and machine self-confidence for enhanced trust in
  autonomous systems.
\newblock In \emph{{RSS} 2016 Workshop on Social Trust in Autonomous Systems}.
  qav.comlab.ox.ac.uk.

\bibitem[{Bar-Shalom, Rong~Li, and Kirubarajan(2001)}]{Bar-Shalom2001}
Bar-Shalom, Y.; Rong~Li, X.; and Kirubarajan, T. 2001.
\newblock \emph{Estimation with Applications to Tracking and Navigation: Theory
  Algorithms and Software}, volume~9.
\newblock Wiley.

\bibitem[{Conlon et~al.(2022)Conlon, Acharya, McGinley, Slack, D'Alonzo,
  Hebert, Reale, Frew, Russell, and Ahmed}]{ConlonSciTech2022}
Conlon, N.; Acharya, A.; McGinley, J.; Slack, C.~A., Trevor~Hirst; D'Alonzo,
  M.; Hebert, M.~R.; Reale, C.; Frew, E.~W.; Russell, R.; and Ahmed, N.~R.
  2022.
\newblock Generalizing Competency Self-Assessment for Autonomous Vehicles Using
  Deep Reinforcement Learning.
\newblock In \emph{2022 AIAA SciTech Forum}.

\bibitem[{Gautam, Crandall, and Goodrich(2021)}]{Gautam2021}
Gautam, A.; Crandall, J.~W.; and Goodrich, M.~A. 2021.
\newblock Self-assessment of Proficiency of Intelligent Systems: Challenges and
  Opportunities.
\newblock In \emph{Advances in Human Factors in Robots, Drones and Unmanned
  Systems}, 108--113. Springer International Publishing.

\bibitem[{Hutchins et~al.(2015)Hutchins, Cummings, Draper, and
  Hughes}]{Hutchins2015}
Hutchins, A.~R.; Cummings, M.~L.; Draper, M.; and Hughes, T. 2015.
\newblock Representing Autonomous Systems' {Self-Confidence} through Competency
  Boundaries.
\newblock In \emph{Proceedings of the Human Factors and Ergonomics Society
  Annual Meeting}, volume~59, 279--283. SAGE Publications.

\bibitem[{Israelsen et~al.(2020)Israelsen, Ahmed, Frew, Lawrence, and
  Argrow}]{Israelsen2020-xQ}
Israelsen, B.; Ahmed, N.; Frew, E.; Lawrence, D.; and Argrow, B. 2020.
\newblock Machine Self-confidence in Autonomous Systems via Meta-analysis of
  Decision Processes.
\newblock In \emph{Advances in Artificial Intelligence, Software and Systems
  Engineering}, 213--223. Springer International Publishing.

\bibitem[{Israelsen(2019)}]{Israelsen2019-thesis}
Israelsen, B.~W. 2019.
\newblock \emph{Algorithmic Assurances and {Self-Assessment} of Competency
  Boundaries in Autonomous Systems}.
\newblock Ph.D. thesis, University of Colorado at Boulder.

\bibitem[{Israelsen and Ahmed(2019)}]{Israelsen2019-survey}
Israelsen, B.~W.; and Ahmed, N.~R. 2019.
\newblock {``Dave...I} Can Assure You ...That It's Going to Be All Right ...''
  A Definition, Case for, and Survey of Algorithmic Assurances in
  {Human-Autonomy} Trust Relationships.
\newblock \emph{ACM Comput. Surv.}, 51(6): 113:1--113:37.

\bibitem[{Kaipa, Kankanhalli-Nagendra, and Gupta(2015)}]{Kaipa2015-hy}
Kaipa, K.~N.; Kankanhalli-Nagendra, A.~S.; and Gupta, S.~K. 2015.
\newblock Toward Estimating Task Execution Confidence for Robotic {Bin-Picking}
  Applications.
\newblock In \emph{2015 {AAAI} Fall Symposium Series}.

\bibitem[{Kochenderfer(2015)}]{Kochenderfer2015}
Kochenderfer, M.~J. 2015.
\newblock \emph{Decision Making Under Uncertainty: Theory and Application}.
\newblock MIT Press.

\bibitem[{Kong(2006)}]{kong2006upm}
Kong, L. 2006.
\newblock The UPM/LPM Framework on Portfolio Performance Measurement and
  Optimization.
\newblock Technical Report 2006:10, Uppsala University, Department of
  Mathematics.

\bibitem[{Kuter and Miller(2015)}]{Kuter2015}
Kuter, U.; and Miller, C. 2015.
\newblock Computational Mechanisms to Support Reporting of Self Confidence of
  {Automated/Autonomous} Systems.
\newblock In \emph{2015 {AAAI} Fall Symposium Series}.

\bibitem[{Leyton-Brown, Nudelman, and Shoham(2009)}]{Leyton-Brown2009}
Leyton-Brown, K.; Nudelman, E.; and Shoham, Y. 2009.
\newblock Empirical Hardness Models: Methodology and a Case Study on
  Combinatorial Auctions.
\newblock \emph{J. ACM}, 56(4): 22:1--22:52.

\bibitem[{Miller(2014)}]{Miller2014-av}
Miller, C.~A. 2014.
\newblock Delegation and Transparency: Coordinating Interactions So Information
  Exchange Is No Surprise.
\newblock In \emph{Virtual, Augmented and Mixed Reality. Designing and
  Developing Virtual and Augmented Environments}, 191--202. Springer
  International Publishing.

\bibitem[{Muir(1994)}]{Muir1994-ow}
Muir, B.~M. 1994.
\newblock Trust in automation: Part I. Theoretical issues in the study of trust
  and human intervention in automated systems.
\newblock \emph{Ergonomics}, 37(11): 1905--1922.

\bibitem[{Sweet et~al.(2016)Sweet, Ahmed, Kuter, and Miller}]{Sweet2016}
Sweet, N.; Ahmed, N.~R.; Kuter, U.; and Miller, C. 2016.
\newblock Towards {Self-Confidence} in Autonomous Systems.
\newblock In \emph{{AIAA} Infotech @ Aerospace}, 1651.

\bibitem[{Zagorecki, Kozniewski, and Druzdzel(2015)}]{Zagorecki2015}
Zagorecki, A.; Kozniewski, M.; and Druzdzel, M. 2015.
\newblock An Approximation of Surprise Index as a Measure of Confidence.
\newblock In \emph{2015 {AAAI} Fall Symposium Series}.

\end{thebibliography}


\end{document}